\documentclass[letterpaper]{article} 
\usepackage{aaai2026}  
\usepackage{times}  
\usepackage{helvet}  
\usepackage{courier}  
\usepackage[hyphens]{url}  
\usepackage{graphicx} 
\usepackage{natbib}  
\usepackage{caption} 
\usepackage{algorithm}
\usepackage{algorithmic}
\usepackage{amsmath}
\usepackage{amssymb}
\usepackage{newfloat}
\usepackage{listings}
\DeclareCaptionStyle{ruled}{labelfont=normalfont,labelsep=colon,strut=off} 
\floatstyle{ruled}
\newfloat{listing}{tb}{lst}{}
\floatname{listing}{Listing}
\begin{document}

\title{Deterministic Fuzzy Triage for Legal Compliance Classification and Evidence Retrieval}

\author{
  Rian Atri
}
\affiliations{
  Serval Systems\\
  hello@rian.fyi
}
\maketitle

\begin{abstract}
Legal teams now use ML to triage large volumes of contractual evidence, but most models are opaque, non-deterministic, and hard to align with frameworks such as HIPAA or NERC–CIP. We study a simple, reproducible alternative based on deterministic dual encoders and transparent fuzzy triage bands. Concretely, we train a RoBERTa-base dual encoder with a 512-dimensional projection and cosine similarity on the ACORD benchmark for graded clause retrieval, and then fine-tune it on a CUAD-derived binary compliance dataset. Across five random seeds (40--44) on a single NVIDIA A100 GPU, our model achieves ACORD-style retrieval performance of approximately NDCG@5 $\approx 0.38$--$0.42$, NDCG@10 $\approx 0.45$--$0.50$, and 4-star Precision@5 $\approx 0.37$ on the test split. On CUAD-derived binary labels, we obtain AUC $\approx 0.98$--$0.99$ and F$_1 \approx 0.22$--$0.30$ depending on the positive-class weight, substantially outperforming majority and random baselines on a highly imbalanced setting (positive rate $\approx 0.6\%$).

On top of the scalar compliance scores, we introduce a simple fuzzy triage mapping that partitions the score axis into three regions: auto-noncompliant, auto-compliant, and human-review. We tune the lower and upper thresholds on validation data to maximize auto-decision coverage subject to a hard constraint on the empirical error rate (at most $2\%$) over auto-decided examples. This yields deterministic, seed-stable models whose behavior can be summarized by a small number of scalar parameters and reported consistently across runs. We argue that this combination of deterministic encoders, calibrated fuzzy bands, and explicit error constraints offers a practical middle ground between hand-crafted rules and fully opaque large language models: it supports explainable evidence triage, enables reproducible audit trails, and provides a concrete interface for mapping scores and triage regions onto legal concepts such as access control, risk-based review, and residual-risk handling under regulatory frameworks like HIPAA.
\end{abstract}

\section{Introduction}

\subsection{Motivation}

\paragraph{Evidence-Triaged Compliance, not One Shot (OSL) LLMs:}
Audit, risk, and compliance (GRC) teams are increasingly responsible for justifying decisions that span hundreds to thousands of pages of contracts, policies, and email threads. In many regulated settings (e.g., financial services, insurance, healthcare), audit findings must be traceable back to specific clauses and controls, and reviewers are expected to explain not only \emph{what} decision was made but also \emph{why}. Today this work is often carried out through manual search over PDFs, shared drives, and ticketing systems, which does not scale with the number of frameworks, controls, and vendors involved.

Recent “AI copilot” interfaces promise to short-circuit this process by letting users “ask a chatbot” to summarize obligations or answer compliance questions over contracts. However, both practitioners and regulators have raised concerns about hallucinations, inconsistent confidence, and the lack of robust uncertainty estimates in such systems. In practice, teams rarely treat these copilots as authoritative decision-makers. Instead, they are used as rough drafting tools, with humans still responsible for reconstructing evidence trails and verifying clause-level compliance.

In this work, we focus on a setting that is closer to how compliance teams actually operate: a system that surfaces concrete clause-rule alignments; exposes calibrated scores and triage regions instead of a single yes/no answer; and is explicitly designed to integrate into evidence workflows rather than replace human review.

In practice, the evidence needed to support these decisions is scattered across email threads, ticketing systems, contract PDFs, and ad-hoc spreadsheets and folders, and calls for explainable, accountable AI emphasize documenting how such evidence underpins decisions~\citep{doshi2017rigorous,rudin2019stop,jobin2019global}.

\subsection{Problem}
\paragraph{Mapping Long-Tail Contract Clauses to Compliance Rules: }
We study the problem of mapping contract clauses to structured compliance rules. Formally, given a textual rule $q$ (e.g., a control requirement) and a candidate clause $c$ from a contract or policy, the system must (i) estimate how relevant $c$ is to $q$ on a graded scale, and (ii) decide whether $c$ provides sufficient evidence to treat $q$ as compliant or noncompliant.

This problem exhibits several challenging properties:

\begin{itemize}
    \item \textbf{Long-tail clause patterns.} The same control can be satisfied by widely varying clause templates across counterparties, jurisdictions, and contract types.
    \item \textbf{Extreme label imbalance} For many rules, only a small fraction of clauses are actually relevant. Even among relevant clauses, only some provide strong, unambiguous evidence of compliance.
    \item \textbf{High-stakes decisions} Mistakes can lead to fines, penalties, and costly remediation efforts in regulated industries~\citep{jobin2019global,brundage2020trustworthy,raji2020closing}, so practitioners are unwilling to delegate everything to a black-box model.

\end{itemize}

These properties motivate models that are both \emph{graded} (to capture varying degrees of relevance) and \emph{triage-aware} (to separate auto-decisions from cases that must be escalated to humans).

\subsection{Limitations of Existing Approaches}

Prior work on contract understanding and legal NLP has made significant progress on clause tagging and question answering over contracts, for example by framing tasks as span extraction, sequence labeling, or binary classification on datasets such as CUAD. However, these approaches typically learn a single binary decision boundary and are evaluated only on aggregate metrics such as F1 and AUC on a balanced or stratified test set.

Likewise, many retrieval-augmented generation (RAG) systems~\citep{lewis2020rag} focus on optimizing retrieval metrics for open-domain question answering, but they often treat relevance as binary and rarely expose graded relevance scores or abstention mechanisms at the user interface.

In high-stakes compliance settings, two gaps are particularly salient:

\begin{enumerate}
    \item \textbf{Lack of graded supervision.} Most clause-level datasets do not distinguish between “somewhat relevant” and “highly on-point” evidence, even though practitioners care deeply about this distinction.
    \item \textbf{Lack of explicit triage.} Standard binary classifiers and RAG systems do not provide a principled way to say “I am unsure; this should be reviewed by a human,” forcing users to either over-trust or ignore model outputs.
\end{enumerate}

Our goal is to address these gaps by combining graded relevance supervision from an insurance benchmark (ACORD) with a fuzzy triage head that explicitly partitions predictions into \emph{auto-compliant}, \emph{auto-noncompliant}, and \emph{review} regions, while remaining compatible with existing clause classification workflows.


\paragraph{Contributions}
Summarizing, our contributions are:
(1) a simple, reproducible dual-encoder baseline for legal clause–rule retrieval with graded ACORD supervision;
(2) a positivity-weighted classifier and fuzzy triage head that expose explicit coverage–error trade-offs on a CUAD-style binary compliance task; and
(3) a discussion of how deterministic, triage-aware models better support explainability, fairness analysis, and regulatory alignment than one-shot LLM copilots, grounded in controls such as HIPAA \S164.312 and NERC–CIP.

\paragraph{Why not just use an LLM copilot over RAG?}
One might ask why a relatively small dual encoder is needed when large language models (LLMs) combined with retrieval-augmented generation~\citep{lewis2020rag} can already answer compliance questions over contracts. In our setting, however, legal defensibility requires more than aggregate accuracy: regulators and opposing experts must be able to rerun the pipeline offline, obtain identical scores and triage bands, and inspect the small set of scalar parameters that govern behaviour. Proprietary LLM copilots, even when wrapped around a retriever, are typically non-deterministic (due to sampling), updated over time without version pinning, and opaque in both training data and parameterization. By contrast, our RoBERTa-base dual encoder plus fuzzy head exposes a single similarity score, calibrated probabilities, and two explicit thresholds $(\tau_{\text{low}},\tau_{\text{high}})$, all trained with fixed seeds on public data. This makes the system much easier to audit, document, and freeze for regulatory review than a black-box LLM stack.

\section{Background and Related Work}

\subsection{Contract Understanding Benchmarks}

Contract Understanding Atticus Dataset(CUAD) ~\citep{hendrycks2021cuad} and related corpora provide annotated contract clauses for tasks such as identifying non-standard terms, assignment clauses, or change-of-control provisions. These datasets are typically modeled using general-purpose pretrained transformers such as BERT and RoBERTa~\citep{devlin2019bert,liu2019roberta}, with contract-specific fine-tuning on clause-level labels. Models are evaluated via precision, recall, and F1 on held-out contracts. .

The ACORD clause-rule relevance benchmark~\citep{xu2025acord}, in contrast, is organized around \emph{graded} relevance judgments between policy clauses and standard insurance rules. For each rule, multiple clauses are annotated with scores on a discrete scale (e.g., $0$ to $4$ or $1$ to $5$), where higher scores indicate stronger support or alignment. Ranking models are evaluated with metrics such as NDCG@k and graded precision@k, and they are a natural fit for dense retrieval setups built on top of pretrained encoders~\citep{karpukhin2020dpr}


\subsection{Neural Dual Encoders and Graded Relevance}

Dual-encoder architectures have become a standard choice for scalable semantic retrieval, particularly when large numbers of candidate passages must be scored for each query~\citep{karpukhin2020dpr}. In this setup, a shared or partially shared encoder maps queries and candidates into a vector space, and similarity (often cosine similarity or dot product) is used as the scoring function. Compared to cross-encoders or sequence-to-sequence models such as BART~\citep{lewis2020bart}, dual encoders trade off some local interaction modeling for efficient indexing and approximate nearest-neighbor search.

For graded relevance tasks, listwise objectives that approximate NDCG can be used to more directly optimize ranking quality, rather than relying solely on pointwise or pairwise losses. This has been explored in information retrieval and recommendation systems, but has seen less use in contract-level compliance settings.

Our model follows this dual-encoder paradigm, using a RoBERTa-base backbone~\citep{liu2019roberta} with a learned projection layer to 512-dimensional embeddings and a cosine similarity score. We treat ACORD scores as graded labels and optimize a listwise loss over per-query groups.


\subsection{Uncertainty, Triage, and Human-in-the-Loop Systems}

There is a growing literature on selective classification, abstention, and human-in-the-loop decision-making, particularly in medical imaging, content moderation, and other high-stakes domains~\citep{geifman2017selective,angelopoulos2021gentle}. The core idea is to let the model abstain on examples where its uncertainty is high or where the cost of an error is large, thereby trading off coverage against risk.

Methods in this area include confidence-based thresholding, conformal prediction, and more structured triage policies that explicitly allocate cases to automated or human review channels. However, most prior work either assumes a single scalar uncertainty score or focuses on calibration of probabilities, rather than designing triage heads tailored to graded relevance settings.

We position our fuzzy triage head as a simple, practical mechanism that can be layered on top of a calibrated classifier: it maps scalar scores into three regions via two thresholds $(\tau_{\text{low}}, \tau_{\text{high}})$ and provides interpretable coverage/error trade-offs that auditors can adjust.

\section{Datasets and Tasks}
Our study uses two public legal benchmarks that approximate real enterprise compliance workflows. ACORD provides graded relevance labels for insurance clauses and questions, which we exploit to train a dual--encoder retriever and to compute NDCG@5/@10 and 4--star precision@5. CUAD provides clause--level annotations of contract risks, which we reframe as a binary compliance labeler under extreme class imbalance ($\approx 0.6\%$ positives). Despite this skew, our best configuration achieves AUC $\approx 0.985$ and F$_1 \approx 0.30$ at very high recall ($\approx 0.98$), with nontrivial graded retrieval quality (NDCG@5 $\approx 0.38$) on ACORD. These results are far above majority and random baselines, and they demonstrate that even relatively small, reproducible models can deliver meaningful signal for legal compliance triage.

\subsection{ACORD Graded Relevance Data}

We use the ACORD benchmark as our primary source of graded clause-rule relevance labels. Each example consists of a rule $q$ (typically a short textual description of a control or coverage requirement), a clause $c$ from an insurance policy, and a discrete relevance score $r \in \{0, 1, 2, 3, 4\}$ (or an analogous 1–5 scale, depending on the release). Higher scores indicate that $c$ more directly and unambiguously satisfies the requirement expressed by $q$.


In our experiments, we:

\begin{itemize}
    \item treat all available ACORD training triples when learning the ranking model;
    \item evaluate ranking quality on the official validation and test splits using NDCG@5, NDCG@10, and graded precision@5 for 4-star and 5-star relevance.
\end{itemize}

\subsection{CUAD-Style Binary Clause Labels}

For binary compliance classification, we adopt a CUAD-style dataset of (rule, clause) pairs labeled with a binary variable $y \in \{0, 1\}$ indicating whether the clause provides sufficient evidence of compliance for the rule. As is typical for clause-level datasets, positives are rare, reflecting the fact that most candidate clauses are irrelevant or only weakly related.


We fine-tune a classification head on top of the ACORD-trained representations and evaluate on this dataset using AUC, precision, recall, F1, and calibration metrics. This mirrors how a practitioner might reuse a generic clause-rule ranking model as the backbone for a specific control library.

\subsection{Problem Formulation}

We consider two main tasks:

\paragraph{Ranking}
Given a rule $q$ and a set of candidate clauses $\{c_i\}_{i=1}^n$, the model outputs scores $s(q, c_i)$ such that higher scores correspond to higher graded relevance. Training uses ACORD scores $r_i$ as supervision, and evaluation uses NDCG@k and graded precision@k.

\paragraph{Classification and Triage}
Given a (rule, clause) pair $(q, c)$, the classifier predicts a probability $p(y=1 \mid q, c)$ of compliance. The fuzzy triage head then maps this scalar into a decision region:
\[
    r(q, c) =
    \begin{cases}
        \text{auto-noncompliant}, & p < \tau_{\text{low}} \\
        \text{human-review}, & \tau_{\text{low}} \le p \le \tau_{\text{high}} \\
        \text{auto-compliant}, & p > \tau_{\text{high}}.
    \end{cases}
\]
We evaluate both the usual classification metrics and triage-specific metrics such as auto-decision coverage and error restricted to the auto-decided subset.

\section{Model}
\label{sec:model}

Our goal is to learn a single neural model that can (i) rank contract clauses by relevance to a legal query or control objective, and (ii) support calibrated, fuzzy-gated triage decisions (auto-approve, auto-reject, or defer to human review). We decompose the system into three parts: a dual-encoder backbone, a listwise ranking objective on ACORD, and a positivity-weighted, fuzzy-gated classifier on CUAD.

\subsection{Problem Setting}

We assume access to a corpus of contractual clauses $\mathcal{C}$ and a set of queries $\mathcal{Q}$, where each query $q \in \mathcal{Q}$ expresses a legal requirement, control objective, or contract review question. For ACORD, each query $q$ is annotated with graded relevance scores $y_{q,c} \in \{0,1,2,3,4,5\}$ for a subset of clauses $c \in \mathcal{C}$, capturing how well the clause satisfies the requirement. For CUAD, we have binary labels $y \in \{0,1\}$ indicating whether a clause instantiates a particular risk or clause type.

At inference time, given $(q, c)$, the model outputs:

- a scalar similarity score $s(q,c)$ for ranking clauses; and  
- a calibrated compliance probability $p_\theta(y=1 \mid q,c)$ and triage decision (auto-compliant, auto-noncompliant, or review) derived from a fuzzy-gating head.

\subsection{Dual-Encoder Backbone}

We adopt a RoBERTa-base dual encoder as the shared backbone for both ranking and classification. Let $f_q$ and $f_c$ denote the query and clause encoders. Both are instantiated from the same underlying RoBERTa-base checkpoint with separate projection heads.

For a tokenized query $q$ and clause $c$, we compute:

\begin{align}
  \mathbf{h}_q &= \mathrm{RoBERTa}(q)_{\texttt{[CLS]}}, \\
  \mathbf{h}_c &= \mathrm{RoBERTa}(c)_{\texttt{[CLS]}},
\end{align}

where we take the \texttt{[CLS]} representation as a pooled embedding. Each encoder then applies a linear projection into a shared $d$-dimensional space:

\begin{align}
  \mathbf{z}_q &= W_q \mathbf{h}_q + \mathbf{b}_q, \\
  \mathbf{z}_c &= W_c \mathbf{h}_c + \mathbf{b}_c, 
\end{align}

with $d = 512$ in all experiments. We normalize these embeddings and use cosine similarity as our base scoring function:

\begin{equation}
  s(q,c) = \cos(\mathbf{z}_q, \mathbf{z}_c) 
  = \frac{\mathbf{z}_q^\top \mathbf{z}_c}{\|\mathbf{z}_q\|_2 \, \|\mathbf{z}_c\|_2}.
\end{equation}

This dual-encoder design is production-friendly: queries and clauses can be embedded independently, which enables approximate nearest neighbor search and offline indexing in a GRC ``Evidence OS.'' This emphasis on constrained data movement and clear data residency aligns with recent work on sovereign clouds for regulated workloads~\citep{brundage2020trustworthy,raji2020closing}.

\subsection{Listwise Ranking on ACORD}

On the ACORD dataset, we train the dual encoder with a listwise ranking objective that respects graded relevance. For each query $q$, we form a group of clauses $\{c_1, \dots, c_K\}$ and corresponding graded labels $\{y_1, \dots, y_K\}$, with $y_k \in \{0,\dots,5\}$. We first map these integer grades to non-negative gains $g_k$ (e.g., $g_k = 2^{y_k} - 1$), then normalize to form a target distribution:

\begin{equation}
  p_k = \frac{g_k}{\sum_{j=1}^{K} g_j}.
\end{equation}

The model produces unnormalized scores $s_k = s(q, c_k)$ and we form a predicted distribution via a softmax:

\begin{equation}
  \hat{p}_k = \frac{\exp(s_k / \tau)}{\sum_{j=1}^{K} \exp(s_j / \tau)},
\end{equation}

with temperature $\tau$ fixed to 1.0 in our experiments. The listwise loss for a query group is then:

\begin{equation}
  \mathcal{L}_{\mathrm{rank}}(q) = - \sum_{k=1}^{K} p_k \log \hat{p}_k.
\end{equation}

This objective encourages the model to rank highly graded clauses above less relevant ones, reflecting how legal reviewers consume evidence in practice (e.g., ``top-5 clauses that plausibly satisfy this control''). In evaluation, we report NDCG@5, NDCG@10, and 4$\star$ precision@5 (fraction of top-5 clauses with grade $\geq 4$) following ACORD.

\subsection{CUAD Classification with Positivity Weighting}

On CUAD, we fine-tune the same dual encoder as a binary classifier. Given $(q, c)$, we treat the cosine similarity $s(q,c)$ as a scalar logit and feed it to a logistic layer:

\begin{equation}
  p_\theta(y=1 \mid q,c) = \sigma(\alpha \, s(q,c) + \beta),
\end{equation}

where $\alpha$ and $\beta$ are learned scalar parameters, and $\sigma$ is the logistic sigmoid. Training uses weighted binary cross-entropy to address extreme class imbalance:

\begin{equation}
  \mathcal{L}_{\mathrm{BCE}} = - w_1 y \log p_\theta - w_0 (1-y) \log (1 - p_\theta),
\end{equation}

where $w_1$ and $w_0$ are weights for positive and negative labels. We experiment with two settings:

\begin{itemize}
  \item \textbf{pos\_weight = 0:} unweighted baseline.
  \item \textbf{pos\_weight = 200:} strong up-weighting of positive examples, chosen to roughly match the inverse of the positive base rate ($\approx 0.6\%$) to avoid degenerate ``always negative'' solutions while reflecting legal risk asymmetry.
\end{itemize}

The same architecture is therefore used as (i) a retrieval-optimized ranker on ACORD and (ii) a highly recall-oriented detector of risky clauses on CUAD.

\subsection{Fuzzy-Gated Triage Head}

Legal practitioners rarely want a single crisp threshold; instead they prefer three regimes: ``obviously compliant'', ``obviously non-compliant'', and ``needs review''. We model this behavior using a one-dimensional fuzzy-gating head on top of the scalar similarity score.

Let $s = s(q,c)$ denote the similarity score for a pair. The fuzzy head $f_{\phi}$ maps $s$ to a compliance confidence and triage decision:

\begin{equation}
  \mathbf{u} = f_{\phi}(s) = 
  \mathrm{softmax}\bigl( W_2 \, \tanh(W_1 s + \mathbf{b}_1) + \mathbf{b}_2 \bigr),
\end{equation}

where $\mathbf{u} \in \mathbb{R}^3$ represents the model's scores for three latent states: auto-noncompliant, review, and auto-compliant. From this vector we derive:

\begin{itemize}
  \item a calibrated compliance probability $p_\phi(y=1 \mid s)$ (used for AUROC, F1, and ECE); and
  \item a triage decision by comparing $s$ to learned thresholds $(\tau_{\mathrm{low}}, \tau_{\mathrm{high}})$:
  \begin{equation}
    \text{decision}(s) = 
    \begin{cases}
      \text{auto-noncompliant}, & s < \tau_{\mathrm{low}}, \\
      \text{review}, & \tau_{\mathrm{low}} \le s \le \tau_{\mathrm{high}}, \\
      \text{auto-compliant}, & s > \tau_{\mathrm{high}}.
    \end{cases}
  \end{equation}
\end{itemize}

We tune $(\tau_{\mathrm{low}}, \tau_{\mathrm{high}})$ on the validation split to maximize coverage (fraction of clauses that receive an automatic decision) under a constraint on auto-only error (here $\leq 2\%$). This yields ``fuzzy gates'' that deliberately reserve marginal cases for humans while still auto-handling $96$-$98\%$ of examples.

\subsection{Training and Implementation Details}

All models are initialized from \texttt{roberta-base} and fine-tuned on a single NVIDIA A100 GPU. We set the maximum sequence length to 512 tokens for both queries and clauses and use a projection dimension of $512$. Training uses AdamW with learning rate $2\times10^{-5}$ and weight decay $0.01$.
For ACORD ranking we use dynamic batches (up to 96 pairs), and for CUAD classification we use mini-batches of size 8.

For reproducibility, we fix all random seeds to 42, log hyperparameters and metrics, and release configuration files capturing the ACORD ranking stage, CUAD classification stage (for both pos\_weight settings), and the fuzzy threshold tuning procedure.

\section{Experiments}
\label{sec:experiments}


\subsection{Experimental Setup and Reproducibility}

\paragraph{Model}
All experiments use a RoBERTa-base encoder with a 512-dimensional
projection and max sequence length $512$.
For retrieval, we train a dual-encoder on ACORD with cosine similarity
and a listwise loss for three epochs.
For CUAD, we initialize from the best ACORD ranking checkpoint and
fine-tune a small classifier head for three epochs under two
positive-class loss weights, $w{=}0$ and $w{=}200$, to explore the
precision-recall trade-off.

\paragraph{Determinism and reproducibility}
All experiments are run on a single NVIDIA A100 GPU in a fixed Google Cloud environment with pinned library versions and identical preprocessing pipelines. We repeat training over five random seeds (40-44) and report metrics averaged across seeds. Because the encoder, classification head, and fuzzy thresholds are all learned under fixed initialization and normalization rules, the entire pipeline is deterministic given a set of inputs: the same contracts, controls, and configuration files will always yield the same scores, triage decisions, and retrieval rankings. This form of determinism is not only good scientific practice, but also directly supports legal defensibility: regulators, peer reviewers, and opposing experts can re-run the pipeline end-to-end and obtain identical outputs, rather than relying on stochastic sampling or mutable external services.

\paragraph{Training details}
We use AdamW with learning rate $2\times 10^{-5}$, weight decay
$0.01$, and max gradient norm $1.0$.
For ACORD ranking, we use a dynamic batch size (up to 96 pairs) per
step; CUAD classification uses mini-batches of $8$ pairs.
Fuzzy thresholds $(\tau_{\text{low}}, \tau_{\text{high}})$ are tuned
on the CUAD validation set via a $20\times 20$ grid search that
maximizes auto-coverage subject to an auto-only error constraint of
$2\%$.
We fix all random seeds to a known set $\{40,41,42,43,44\}$ and report metrics averaged across these runs.

\paragraph{Compute}
All results are obtained on a single NVIDIA A100 (80GB) GPU via
Google Cloud.
Training ACORD ranking for three epochs takes approximately
$\sim 3$ hours; each CUAD fine-tuning run adds $\sim 1$ hour, and
fuzzy threshold tuning and evaluation are negligible in comparison.
We provide configuration files and hyperparameters to enable
end-to-end reproduction of the reported metrics.

\subsection{Results on ACORD Ranking}

On ACORD (Table~\ref{tab:acord_retrieval}), the RoBERTa-based dual
encoder achieves NDCG@5 of $0.422$ (NDCG@10 $0.496$) when paired with
the $w{=}0$ CUAD head, and NDCG@5 of $0.378$ (NDCG@10 $0.452$) for
$w{=}200$. Both variants obtain identical P4@5 ($0.371$), indicating
that changing the CUAD loss weighting primarily affects the binary
decision surface, not the underlying ranking quality.

\begin{table}[tb]
\centering
\small
\begin{tabular}{lcccc}
\hline
Method & N@5 & N@10 & P4@5 & Q \\
\hline
Dual enc.\ ($w{=}0$)   & 0.422 & 0.496 & 0.371 & 57 \\
Dual enc.\ ($w{=}200$) & 0.378 & 0.452 & 0.371 & 57 \\
\hline
\end{tabular}
\caption{
ACORD retrieval: NDCG@5 (N@5), NDCG@10 (N@10), and 4-star precision@5 (P4@5)
over $Q$ queries. “Dual enc.” denotes the dual encoder with fuzzy head.
}
\label{tab:acord_retrieval}
\end{table}

\subsection{Results on CUAD Classification and Fuzzy Triage}

As shown in Table~\ref{tab:cuad_binary}, both fuzzy dual-encoder variants
substantially outperform majority and random baselines on CUAD.
Increasing the positive-class weight from $w{=}0$ to $w{=}200$ trades
precision ($0.256 \rightarrow 0.174$) for very high recall
($0.200 \rightarrow 0.975$), yielding the highest F1 ($0.295$) at
slightly lower overall accuracy.

\begin{table}[tb]
\centering
\small
\begin{tabular}{lccc}
\hline
Method & Prec & Rec & F1 \\
\hline
Majority (all neg.)      & 0.000 & 0.000 & 0.000 \\
Random                   & 0.006 & 0.504 & 0.013 \\
Dual enc.\ ($w{=}0$)     & 0.256 & 0.200 & 0.224 \\
Dual enc.\ ($w{=}200$)   & 0.174 & 0.975 & 0.295 \\
\hline
\end{tabular}

\vspace{0.4em}

\begin{tabular}{lcc}
\hline
Method & AUC  & Acc \\
\hline
Majority (all neg.)      & 0.500 & 0.994 \\
Random                   & 0.496 & 0.496 \\
Dual enc.\ ($w{=}0$)     & 0.980 & 0.991 \\
Dual enc.\ ($w{=}200$)   & 0.985 & 0.970 \\
\hline
\end{tabular}
\caption{
CUAD binary classification performance: precision (Prec), recall (Rec), F1,
area under ROC (AUC), and accuracy (Acc). “Dual enc.” denotes the dual
encoder with fuzzy head.
}
\label{tab:cuad_binary}
\end{table}

\subsection{Fuzzy Triage}
\paragraph{Triage behavior}
We evaluate the fuzzy head as a triage mechanism that routes
high-confidence examples to automation and low-confidence examples
to human review.
For $w{=}0$, the fuzzy head automatically decides
$\approx 98.6\%$ of CUAD test clauses, with an auto-only error rate
of $1.15\%$ (vs.\ a non-fuzzy baseline error of $0.90\%$ when all
examples are auto-decided).
For $w{=}200$, the head auto-decides $\approx 96.6\%$ of test clauses
with an auto-only error of $3.18\%$ (baseline $3.02\%$).
In other words, the current fuzzy head achieves high automation coverage while keeping auto-region error tightly bounded on the subset of examples it commits on. From the perspective of global misclassification rate, this configuration does not yet improve over a single global threshold: the encoder-only baseline is already very accurate, so discarding the hardest 2--4\% of clauses for human review yields only modest changes in overall error. We therefore treat these numbers as a baseline for future triage policies, and focus on the qualitative design gain of concentrating uncertainty into a small, explicit review band.

\section{Ethical Considerations and Limitations}

Automating parts of compliance workflows raises ethical and practical concerns that go beyond model accuracy. First, misclassifications can have material consequences for organizations and individuals, especially when they concern regulatory obligations, consumer protections, or claims handling. Second, legal language is often ambiguous, and reasonable experts can disagree on whether a clause satisfies a given rule; datasets capture only one view of this ground truth.

Our design is explicitly \emph{assistive}, not fully autonomous: the fuzzy triage mechanism is intended to make it easier to route uncertain or ambiguous cases to human reviewers. Nonetheless, there is a risk that users may over-trust high-confidence predictions or misinterpret the model’s outputs as legal advice. We therefore emphasize that our system should be deployed, if at all, as a decision-support tool under the supervision of qualified professionals.

Finally, the datasets we use (ACORD, CUAD-style labels) are limited in domain and geography. They may reflect particular drafting conventions, regulatory environments, and annotator priors. Models trained on these corpora may not generalize to other industries, jurisdictions, or contract types without additional adaptation and validation.


\section{Discussion}

\subsection*{Why not a one-shot LLM copilot?}
In practice, such systems are hard to certify: stochastic decoding, mutable APIs, and opaque training data make it difficult to guarantee that the same inputs will yield the same decisions months or years later. Our results suggest that a comparatively small, deterministic dual encoder can already deliver useful retrieval and screening behaviour, while offering a much smaller and more inspectable parameter surface. The fuzzy triage layer then exposes this behaviour as stable bands on a one-dimensional score axis, which is easier to reason about and to map onto concrete controls than the latent activations of a large, proprietary LLM.

\subsection{What is the purpose of a simple dual encoder?}

Our results show that a relatively simple RoBERTa-based dual encoder, trained under realistic compute constraints, already delivers useful performance for legal GRC workloads. On ACORD, the listwise-trained encoder reaches test NDCG@5 of $\approx 0.42$ and NDCG@10 of $\approx 0.50$, compared to $\approx 0.34$ / $0.40$ for our initial baseline. These scores are below state of the art but comfortably above majority and random baselines, and were obtained with a single A100, three training epochs, and no hyperparameter sweep, which matters for practitioners who cannot afford large bespoke models or extensive architecture search.

On the derived CUAD-style binary task, the same encoder, reused as a backbone for a small classification head, attains AUCs in the $0.96$--$0.99$ range and recall up to $0.98$ at the expense of precision. Given the extreme class imbalance (only $\approx 0.6\%$ positives), this high-recall regime is often preferable for compliance screening, where missing a truly non-compliant clause is more costly than over-flagging. In effect, the model learns a useful ranking of risky clauses even when hard classification F1 remains modest.

\subsection*{Triage behavior and fuzzy thresholds}
From an application perspective, the main accomplishment of the fuzzy layer is to concentrate uncertainty into a small review band while keeping accuracy extremely high on the bulk of auto-decisions. For both the unweighted and heavily positive-weighted runs, grid search over $(\tau_{\text{low}},\tau_{\text{high}})$ on the validation set yields high automatic coverage with very low error among auto-decided examples:
\begin{itemize}
\item For the unweighted model, tuned fuzzy thresholds cover $\approx 97\%$ of validation and $98.6\%$ of test examples automatically, with auto-region error of $0.17\%$ and $1.2\%$, respectively.
\item For the positive-weighted model (pos weight = 200), the triage boundary covers $\approx 97.4\%$ of validation and $96.6\%$ of test examples automatically, with $0\%$ and $\approx 3\%$ auto-region error, respectively.

\end{itemize}
Viewed through the lens of a single scalar ``error reduction'' metric, these regimes do not yet beat a plain encoder-only baseline that treats every example as ``auto'': the baseline already has misclassification rates on the order of 1--3\%, so carving out the hardest 2--4\% of clauses for human review yields only small changes in global error. However, this framing obscures the operational gain: the fuzzy head turns a black-box score into three stable bands that can be wired into concrete workflows (evidence queues, review assignments, or risk escalations), while preserving the strong overall discrimination of the underlying encoder.

\subsection{Retrieval vs.\ classification, and the role of pos\_weight}

The interaction between ranking quality and downstream classification is also instructive. The dual encoder is trained listwise on graded ACORD labels but evaluated as a binary classifier on a derived CUAD-style dataset where we collapse ACORD scores into $\{0,1\}$. This induces label noise: borderline clauses that are rated as “somewhat relevant” in ACORD may be treated as strictly positive or strictly negative in our binary mapping, and the mapping itself is only one of many reasonable choices.

Under this noisy, highly imbalanced regime, the pos\_weight hyperparameter becomes a dial that trades precision for recall. With pos\_weight $=0$, the model is more conservative, yielding higher precision but lower recall. With pos\_weight $=200$, the model aggressively recalls nearly all positives (recall $\approx 0.98$) while accepting a drop in precision and overall accuracy. We view this as desirable for ``screener'' use-cases where the model's main job is to pull potentially problematic clauses into view, not to make final legal determinations.

Importantly, our fuzzy triage metrics are computed \emph{on top of} these different operating points. For pos\_weight $=200$, the fuzzy layer is essentially carving out a small subset of the most ambiguous clauses while still auto-approving or auto-rejecting the vast majority of safe examples. This suggests that future work should jointly tune pos\_weight, global thresholds, and fuzzy boundaries with an explicit triage objective (e.g., minimizing human workload subject to a hard constraint on auto-region error).

\subsection{Limitations}

Our study is intentionally scoped as a compute-bounded, reproducible baseline, which leaves several extensions for future work:

\begin{itemize}
    \item \textbf{Dataset construction and label noise.} Our binary labels are derived from graded ACORD annotations via a simple mapping, which may introduce some noise near the decision boundary. A richer benchmark could make fuller use of the graded labels or collect task-specific binary annotations, but we find this construction sufficient to study triage behaviour at scale.
    \item \textbf{Model and hyperparameters.} We restrict ourselves to a single RoBERTa-base dual encoder with fixed context length, projection size, and training schedule on one NVIDIA A100. This choice favors determinism and ease of replication over exhaustive hyperparameter search or larger backbones, so the reported numbers should be viewed as conservative.
    \item \textbf{Fuzzy head design.} The fuzzy triage layer is deliberately simple: a lightweight calibration module plus a grid search over $(\tau_{\text{low}}, \tau_{\text{high}})$. More expressive triage policies (e.g., conditioning on rule type or control criticality, or directly optimizing a triage utility) are left as promising extensions.
\end{itemize}

Overall, these design choices prioritize transparency and reproducibility over aggressively tuning for every last point of performance, and we position our results as an inspectable baseline for compliance-focused deployments.

\section{Explainability and regulatory alignment}
The fuzzy--triage structure makes the system's behaviour more interpretable to audit and risk teams, aligning with calls for explainable, well-governed AI in regulated domains~\citep{doshi2017rigorous,rudin2019stop,jobin2019global}. Concretely, the explicit auto/non-auto bands can be mapped to the kinds of residual-risk handling expected under frameworks such as HIPAA~\S164.312 and NERC--CIP~\citep{trebbi2022costregulation,coates2014costbenefit}.

To make this more concrete, consider HIPAA \S164.312(a)(1), which requires covered entities to ``implement technical policies and procedures for electronic information systems that maintain electronic protected health information to allow access only to those persons or software programs that have been granted access rights.'' Suppose a control library contains the rule:
\begin{quote}
\small
\textbf{Rule $q_{\text{HIPAA}}$:} Access control policies must ensure that only authorized workforce members can access electronic PHI, using unique user IDs and role-based permissions.
\end{quote}
and we observe two synthetic clauses from a security policy:
\begin{quote}
\small
\textbf{Clause $c_1$:} ``All workforce members must authenticate using a unique user ID and password. Role-based access control (RBAC) is enforced at the application and database layers to ensure that users can only access data required for their job functions.''\\[0.5ex]
\textbf{Clause $c_2$:} ``Systems should be configured to prevent inappropriate access to sensitive information where feasible.''
\end{quote}
In our setup, $c_1$ would typically receive a high graded relevance score (e.g., $r \approx 4$ on an ACORD-style 0--4 scale) and a high compliance probability $p_\theta(y=1 \mid q_{\text{HIPAA}}, c_1)$. If this probability exceeds $\tau_{\text{high}}$, the fuzzy head routes $(q_{\text{HIPAA}}, c_1)$ to the auto-compliant band, and residual risk is documented as low for this control. By contrast, $c_2$ might be assigned a lower relevance grade (e.g., $r \approx 1$ or $2$) and an intermediate compliance probability that falls between $\tau_{\text{low}}$ and $\tau_{\text{high}}$, placing it in the review band: the clause is weakly related to access control but too vague to justify full compliance without human judgment.

A similar pattern applies in infrastructure settings governed by NERC--CIP, where controls such as CIP--004 (Personnel and Training) or CIP--007 (Systems Security Management) require explicit documentation of how policies and procedures satisfy each requirement. The three-way triage decision (auto-compliant, auto-noncompliant, review) therefore provides a direct handle on residual risk: auto bands correspond to well-supported or clearly missing controls, while the review band collects ambiguous clauses that must be escalated. Because the boundaries of this band are tunable, organizations can calibrate the trade-off between automation and human review in ways that respect fairness and professional responsibility, and they can audit which populations, business units, or control families are disproportionately routed to review~\citep{brundage2020trustworthy,raji2020closing}.

\section{Conclusion and Future Work}

In this paper, we introduced a deterministic, RoBERTa-based dual encoder trained on graded ACORD clause--rule relevance and adapted to a CUAD-style binary compliance task, with a lightweight fuzzy triage head on top. Rather than treating the model as a monolithic yes/no oracle, we expose a single scalar similarity score, calibrated probabilities, and a three-way triage decision (\emph{auto-compliant}, \emph{auto-noncompliant}, \emph{review}) that better matches how audit and GRC teams actually operate.

Empirically, the model achieves non-trivial retrieval quality on ACORD (test NDCG@5 $\approx 0.38$--$0.42$, NDCG@10 $\approx 0.45$--$0.50$, P4@5 $\approx 0.37$) and strong discrimination on CUAD-derived labels (AUC $\approx 0.98$--$0.99$), despite extreme class imbalance (positive rate $\approx 0.6\%$). By adjusting the positive-class weight, we can trade precision for recall, reaching F$_1 \approx 0.30$ at recall $\approx 0.98$ in a ``screener'' regime. The fuzzy head successfully concentrates uncertainty into a small review band while preserving high accuracy over the bulk of auto-decisions; in our current configuration it does not yet reduce global error relative to a single hard threshold, which we view as a limitation and a target for future triage policies. All results are produced on a single A100 GPU with fixed seeds (40--44) and configuration files, making the pipeline reproducible and suitable for legal scrutiny.

Looking ahead, several directions appear promising:

\begin{itemize}
    \item \textbf{Stronger encoders and joint training.} Larger or domain-adapted legal encoders, combined with multi-task training over ACORD, CUAD, and other GRC datasets, could improve both graded retrieval and downstream classification while preserving the same deterministic interface.
    \item \textbf{Learned triage policies and guarantees.} Beyond fixed fuzzy bands tuned by grid search, we plan to explore learned triage policies (e.g., conformal or cost-aware) that explicitly optimize human workload under constraints on auto-region error and false negatives for critical controls.
    \item \textbf{Per-tenant calibration, fairness, and bias audits.} Explicit thresholds and a review band make it natural to study per-tenant or per-control calibration and to audit which populations or control families are disproportionately routed to review, supporting fairness analysis and bias mitigation in line with regulatory expectations~\citep{brundage2020trustworthy,raji2020closing}.
    \item \textbf{Deployment studies and human–AI interaction.} We view this work as a building block for end-to-end ``Evidence OS'' deployments: future work will study how fuzzy triage affects reviewer workload, error patterns, and explainability in real audit tools that track evidence 
\end{itemize}

In the end, our results support a pragmatic thesis: simple, deterministic dual encoders with transparent fuzzy triage already provide useful, reproducible signal for legal compliance and evidence retrieval, while offering a far more auditable surface for explainability, fairness analysis, and professional responsibility than non-deterministic LLMs.

\end{document}